\documentclass{article}
\usepackage{ijcai24}

\usepackage{times}
\usepackage{soul}
\usepackage{url}
\usepackage[hidelinks]{hyperref}
\usepackage[utf8]{inputenc}
\usepackage[small]{caption}
\usepackage{graphicx}
\usepackage{amsmath}
\usepackage{amsthm}
\usepackage{booktabs}
\usepackage{algorithm}
\usepackage{algorithmic}
\usepackage[switch]{lineno}

\usepackage{graphics}
\usepackage{epstopdf}
\usepackage{epsfig}
\usepackage{array}
\usepackage{amssymb}
\usepackage{mathrsfs}

\usepackage{subfigure} 

\usepackage{multirow}
\usepackage{xcolor}
\usepackage{bm}
\usepackage{eqparbox}
\usepackage{algorithm}  
\usepackage{algorithmic}
\usepackage{graphicx}
\newtheorem{myTheo}{Theorem}

\usepackage{amsfonts}
\usepackage{pifont}
\usepackage{enumerate}
\usepackage{indentfirst}
\usepackage{url}
\usepackage{caption}

\usepackage{ragged2e}
\usepackage{nicefrac}       
\usepackage{microtype}      
\usepackage{times}
\usepackage{enumitem}
\usepackage{wrapfig}
\newtheorem{myMech}{Mechanism}

\title{CNN2GNN: How to Bridge CNN with GNN}

\author{
Ziheng Jiao$^{1,2,3,\dag}$
\and
Hongyuan Zhang$^{1,2,3,\dag}$\and
Xuelong Li$^{3,*}$
\\
\affiliations
$^1$ School of Computer Science, Northwestern Polytechnical University\\
$^2$ School of Artificial Intelligence, Optics and Electronics (iOPEN), Northwestern Polytechnical University\\
$^3$ Institute of Artificial Intelligence (TeleAI), China Telecom Corp Ltd\\
$^{\dag}$ Equal Contribution. $^*$ Corresponding Author. \\
\emails
jzh9830@163.com,
hyzhang98@gmail.com,
li@npwu.edu.cn
}

\begin{document}

\maketitle

\begin{abstract}
Although the convolutional neural network (CNN) has achieved excellent performance in vision tasks by extracting the intra-sample representation, it will take a higher training expense because of stacking numerous convolutional layers. Recently, as the bilinear models, graph neural networks (GNN) have succeeded in exploring the underlying topological relationship among the graph data with a few graph neural layers. Unfortunately, it cannot be directly utilized on non-graph data due to the lack of graph structure and has high inference latency on large-scale scenarios. Inspired by these complementary strengths and weaknesses, \textit{we discuss a natural question, how to bridge these two heterogeneous networks?} In this paper, we propose a novel CNN2GNN framework to unify CNN and GNN together via distillation. Firstly, to break the limitations of GNN, a differentiable sparse graph learning module is designed as the head of networks to dynamically learn the graph for inductive learning. Then, a response-based distillation is introduced to transfer the knowledge from CNN to GNN and bridge these two heterogeneous networks. Notably, due to extracting the intra-sample representation of a single instance and the topological relationship among the datasets simultaneously, the performance of distilled ``boosted'' two-layer GNN on Mini-ImageNet is much higher than CNN containing dozens of layers such as ResNet152.
\end{abstract}

\section{Introduction}\label{sec:introduction}
Convolution neural network (CNN) utilizes the convolution kernel to project the image into the deep space and extract the intra-sample representation such as the color and texture of a single instance, it can obtain great improvements in many image-related tasks \cite{he2016deep}. Notably, since the convolution kernel can be viewed as the variant of linear projection with the parameters sharing \cite{bishop2006pattern}, CNN generally will stack plenty of convolution neural layers to improve the representational ability. However, the large network size may takes the vast resource for storage and optimization. Besides, these linear projections may limit the capacity to extract the multi-type representation, e.g., CNN generally focuses on learning the intra-sample representation but ignores extracting the latent topological relationship among the instance sets \cite{he2022rel}. Although some researchers introduce transferring learning \cite{oquab2014learning}, knowledge distillation \cite{hinton2015distilling}, and network pruning \cite{yu2020easiedge} to achieve the compression of the large CNN, the compressed CNN is still based on linear projection and even the performance will degrade \cite{phuong2019towards}. 

\begin{figure}[t]
	\centering
	\subfigure[CNN: Extract the Deep Intra-Sample Representation]{
		\label{CNN_frame}
		\includegraphics[scale=0.36]{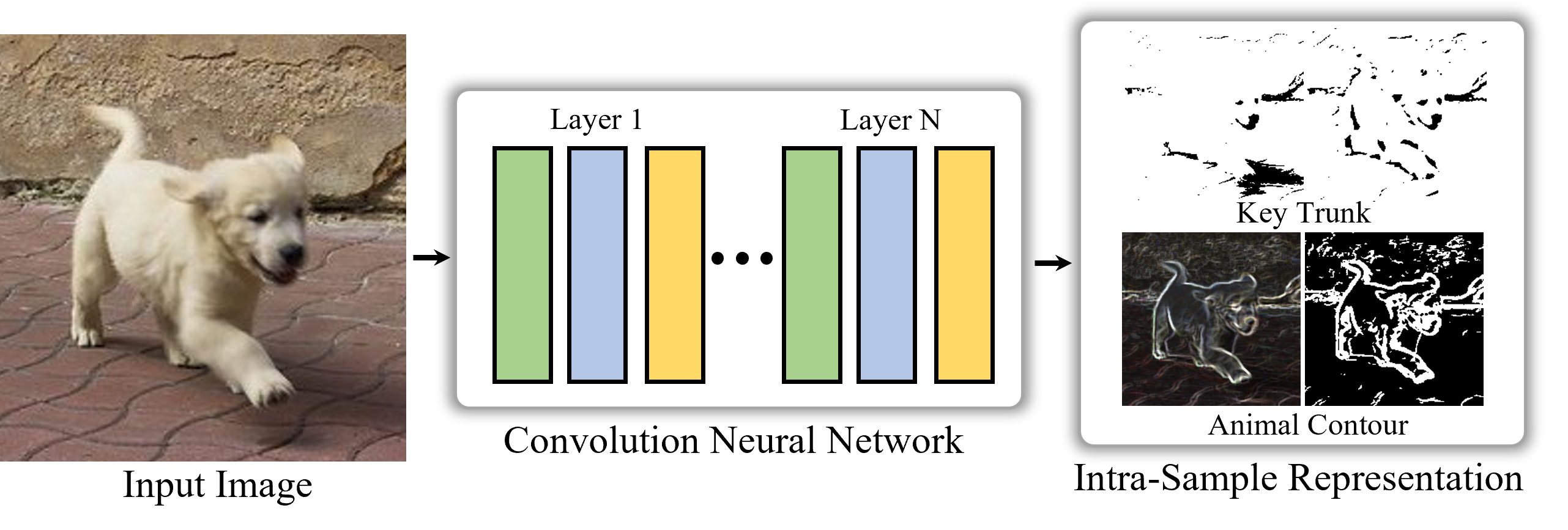}
	}
	\subfigure[GNN: Learn the Underlying Topological Relationship]{
		\label{GNN_frame}
		\includegraphics[scale=0.36]{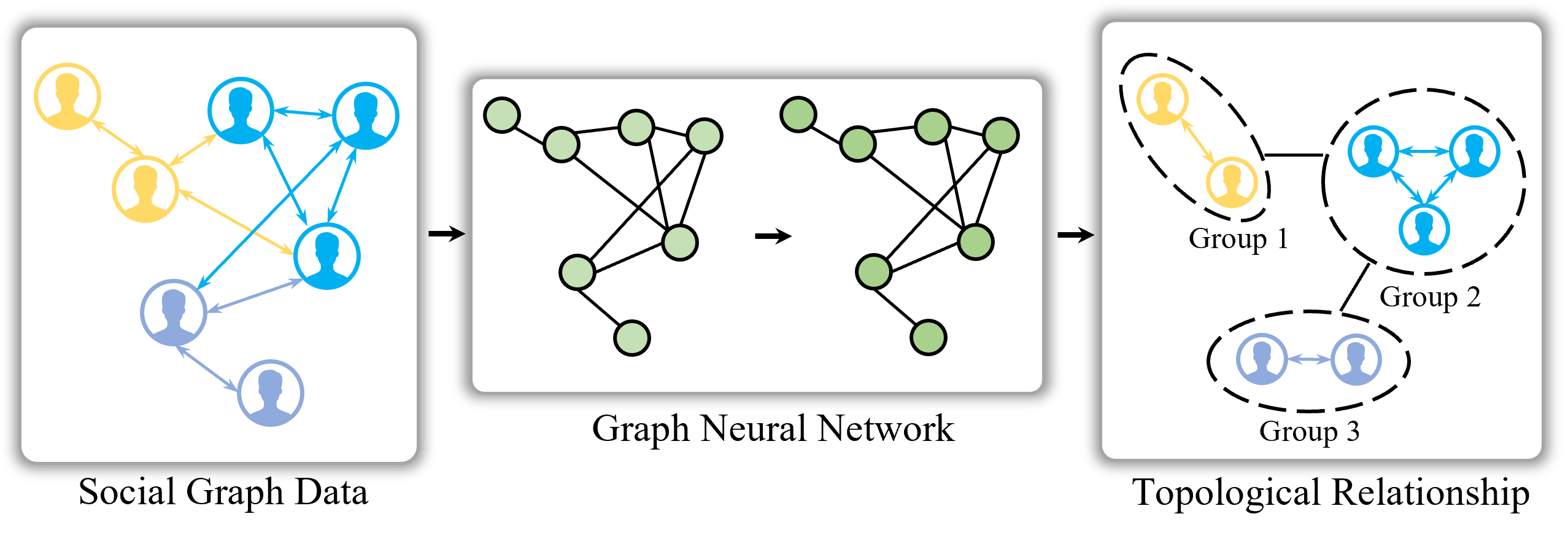}
	}
	\vspace{-3mm}
	\caption{Merits of CNN and GNN. Figure \ref{CNN_frame}: CNN can extract the intra-sample representation such as the trunk and contour of the animal in the image. Figure \ref{GNN_frame}: GNN will learn the explore the relationship among the nodes in the social graph data.}
	\label{intro}
	\vspace{-5mm}	
\end{figure}

\begin{figure*}[t]
	\centering
	\subfigure[Inductive Training of CNN2GNN]{
		\label{framework1}
		\includegraphics[scale=0.43]{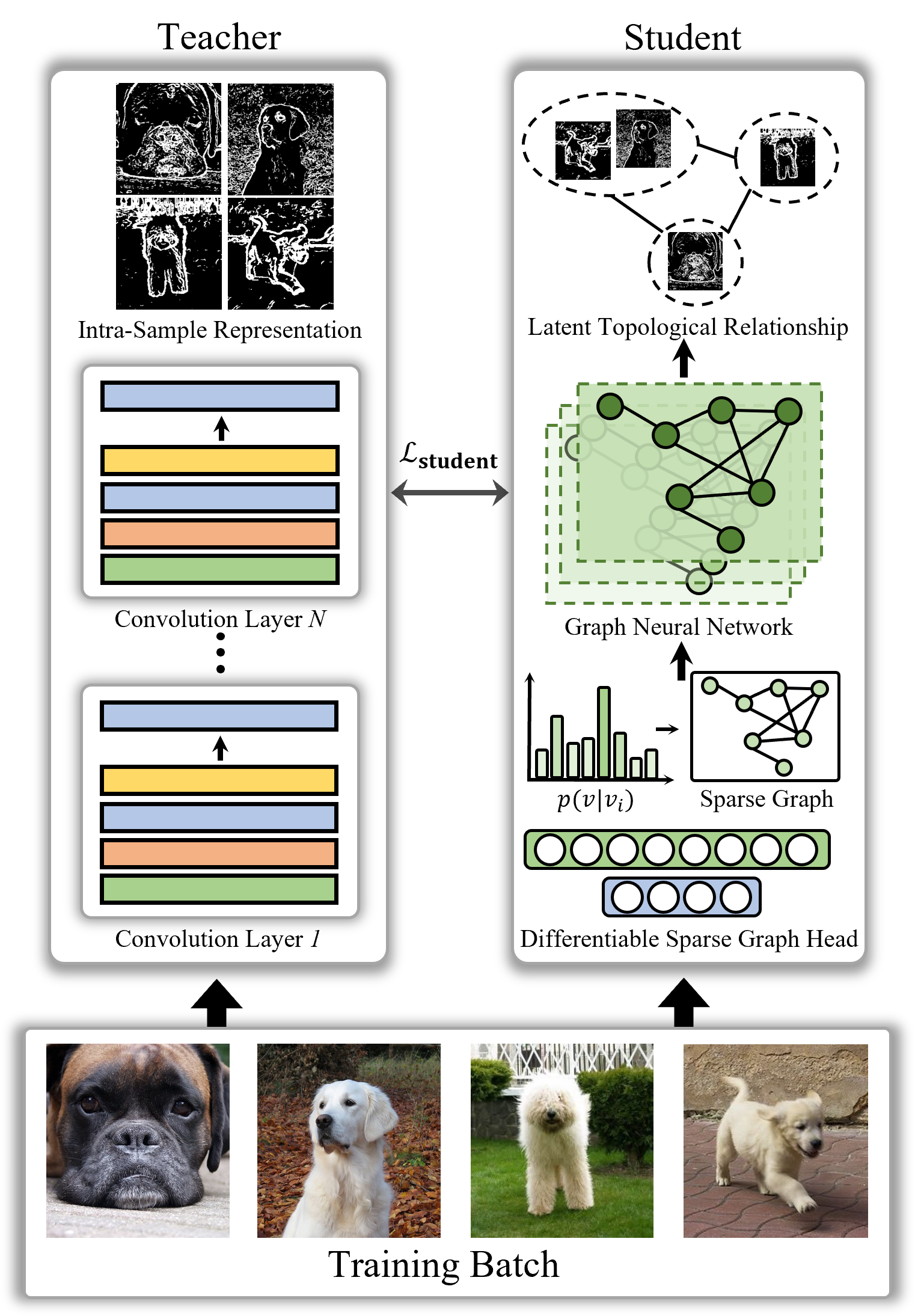}
	}
	\subfigure[Inductive Testing with Mechanism \ref{one_instance}]{
		\label{framework2}
		\includegraphics[scale=0.43]{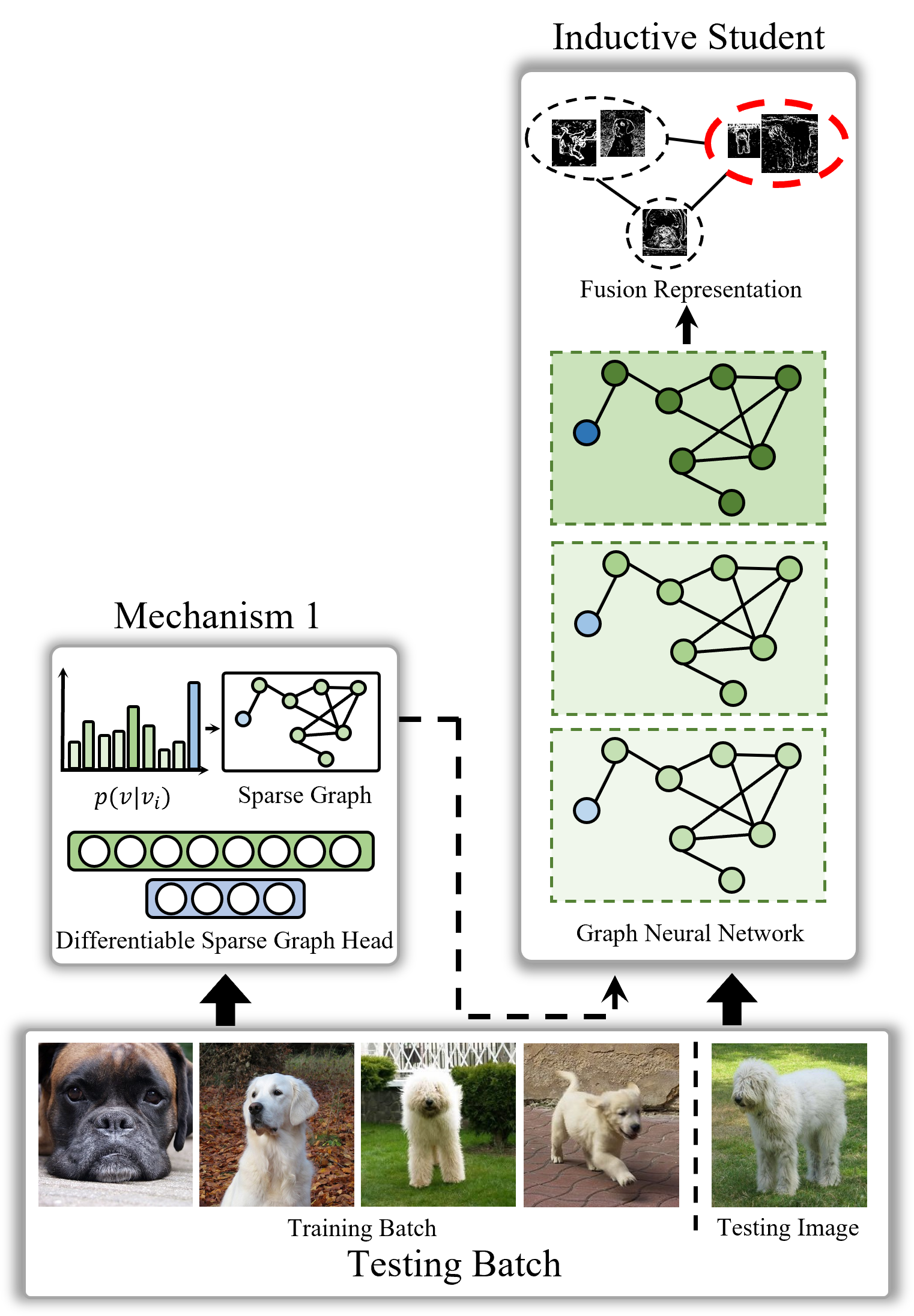}
	}
	\subfigure[Inductive Testing with Mechanism \ref{batch_instance}]{
		\label{framework3}
		\includegraphics[scale=0.43]{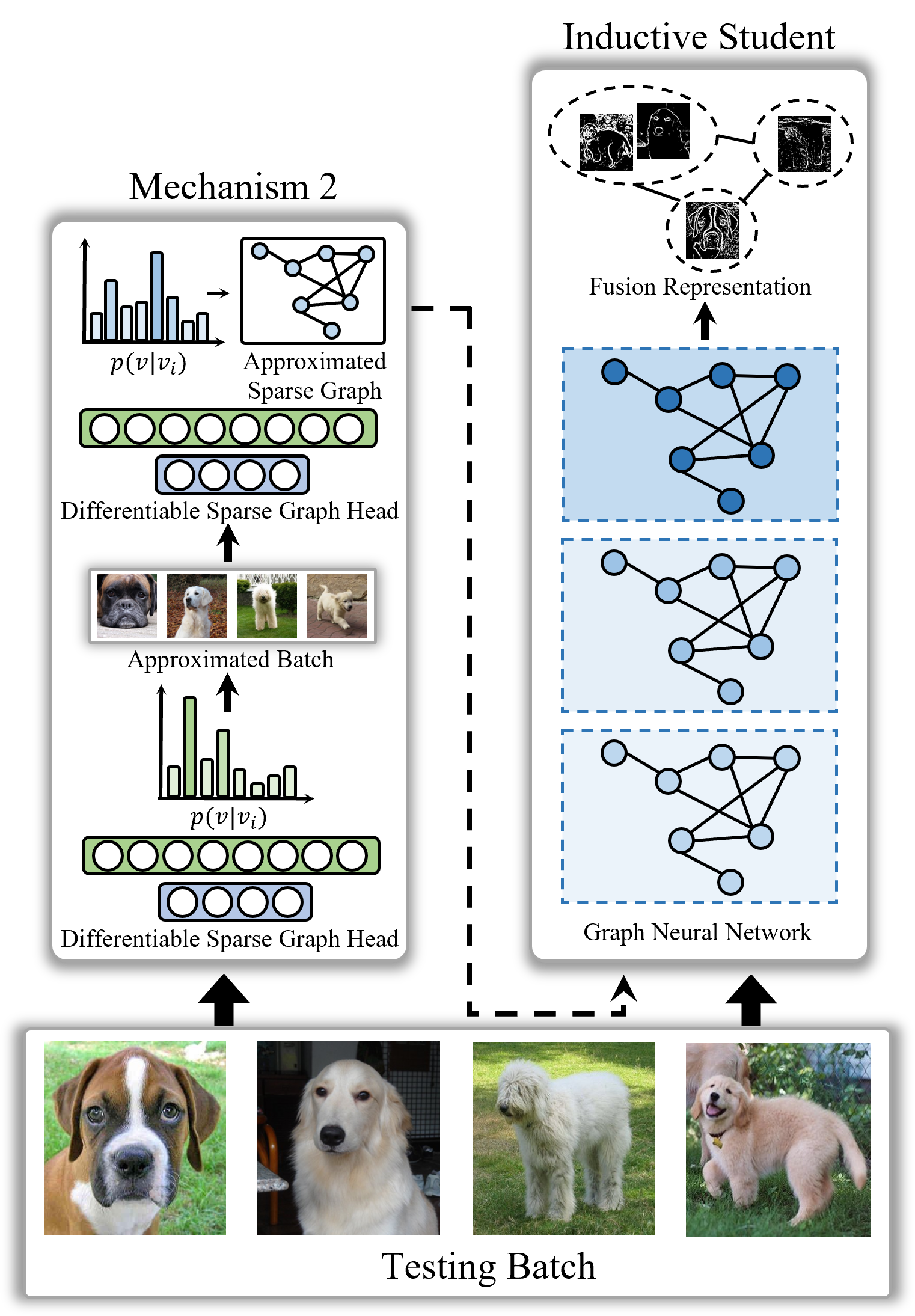}
	}
	\vspace{-3mm}
	\caption{A framework of the proposed CNN2GNN model. Figure \ref{framework1} is the training procedure. CNN teacher and GNN student are utilized to learn the intra-sample representation and the latent topological relationship, respectively. $\mathcal{L}_{\rm student}$ can brige these two heterogeneous networks and transfer knowledge from CNN to GNN. Figure \ref{framework2} and Figure \ref{framework3} are the inductive inference with Mechanism \ref{one_instance} and \ref{batch_instance}, respectively. Among them, Figure \ref{framework2} cascades a test instance with a training batch for prediction. Figure \ref{framework3} selects the most similar sample in the training set to learn an approximated graph structure for evaluating the testing samples batch-by-batch.}
	\label{framework}	
	\vspace{-5mm}
\end{figure*}

Recently, a graph neural network (GNN) has obtained the excellent performance on the graph scenarios such as citation networks, social networks, and biomolecule structure datasets \cite{kipf2016semi}. Compared with linear projection based CNN, GNN is a bilinear model according to \cite{tenenbaum2000separating}. Specifically, it employs two linear factors, projection matrix $\bm W$ and graph-related matrix $\bm P$, to project the data into the deep subspace and aggregate the information from the neighbors. Thus, it can achieve to extract the latent topological relationship among the graph nodes within a few graph neural layers and takes less computation resources than CNN. Meanwhile, it is these two linear factors for projection that make it possible for GNN with the limited layers to learn multi-type information from the data, e.g., extracting the intra-sample representation and latent topological relationship simultaneously on image-related tasks. 


Based on the above discussion, we notice that CNN achieves great performance in the image set by extracting the intra-sample representation as shown in Figure \ref{CNN_frame}. However, due to projecting with a linear factor, CNN generally stacks plenty of neural layers to extract the intra-sample representations, whose optimization will take a vast cost. Meanwhile, although GNN succeeds in extracting the relationship information with a few layers by employing two linear factors for projecting as suggested in Figure \ref{GNN_frame}, the obstacles including graph dependency and high inference latency limit GNN to extend the non-graph data directly. Thus, inspired by these complementary strengths and weaknesses between CNN and GNN, we raise a natural question, \textit{how to bridge these two heterogeneous networks, enjoying the intra-sample representation from CNN and the topological relationship from GNN simultaneously?}

In this work, we find that the response-based heterogeneous distillation can distill the knowledge from CNN to GNN and answer this question. Meanwhile, to eliminate the obstacle between CNN and GNN, we design a differentiable sparse graph learning module as the head in GNN. Due to differentiability, it can be trained with gradient descent and learn a sparse graph for inductive learning. Later, with the assistance of this head and heterogeneous distillation, the distilled ``boosted'' GNN can inductively learn the intra-sample representation of a single instance and the topological relationship among the instance set simultaneously. Notably, the performance of distilled ``boosted'' two-layer GNN on Mini-ImageNet is much higher than CNN containing dozens of layers such as ResNet152. Our core contributions are as follows:
\begin{itemize}[leftmargin=8pt, topsep=0pt, itemsep=0pt]
	\item To eliminate the obstacles between CNN and GNN, a graph head is designed. It can inductively learn the differentiable sparse graph with gradient descent on the non-graph data.
	\item According to response-based distillation, a novel CNN2GNN framework is proposed to distill the knowledge from the large CNN to a tiny GNN and bridge these two heterogeneous networks.
	\item The distilled ``boosted'' GNN can inductively extract the intra-sample representation of a single instance and the topological relationship among the instances simultaneously.
\end{itemize}

\section{Related Work}
\subsection{Graph Neural Network}
Graph neural networks have achieved great performance on graph-structured data such as molecules formed network \cite{duvenaud2015convolutional}. According to spectral graph theory,  \cite{bruna2013spectral} define the parameterized filters in the spectral domain and have been used for graph classification. However, it is time-consuming for these methods to complete the spectral decomposition. Thus, the Chebyshev polynomial \cite{defferrard2016convolutional} and the first-order expansion formed \cite{kipf2016semi} are employed to fit the results of decomposition. Based on this, this formed graph neural network is widely used in graph scenarios \cite{li2022deep}. Meanwhile, since GNN can learn the latent structure information, many researchers attempt to extend it to non-graph datasets  \cite{9606581,zhang2021deep}. Although these methods construct the graph on non-graph data, the obtained is fixed during the training and inference due to the independence between construction and gradient descent. Meanwhile, the inference latency puzzle the GNN applied in the realistic scenario. To overcome these problems and eliminate the obstacle between GNN and CNN, our work designs a graph head to inductively learns a differentiable sparse graph.

\subsection{Knowledge Distillation}
Knowledge distillation mainly transfers the knowledge of a pretrained teacher network into a smaller student network \cite{hinton2015distilling}. According to the categories of transferring knowledge, distillation generally can be classified into three types. The response-based distillation enables the student to directly mimic the response of the decision layer of the teacher network \cite{hinton2015distilling}. 
Then, the feature-based distillation attempt to learn the knowledge in the intermediate layer of the teacher network \cite{zagoruyko2016paying}. 
Furthermore, the relation-based knowledge focuses on learning relationship-level information among the instances \cite{liu2019structured}. However, since the mentioned methods mainly distill the knowledge between the homogenous neural network such as CNN to CNN or GNN to GNN, they mainly focus on transferring the learned knowledge and lack learning unexplored knowledge. In this work, inspired by that the different kinds of neural networks focus on learning different knowledge, we propose a novel heterogeneous distillation strategy to bridge a large CNN and a small GNN. The distilled ``boosted'' GNN can extract the deep intra-sample representation and the topological relationship among the instance simultaneously.

\begin{figure*}[t]
	\centering
	\includegraphics[width=170mm]{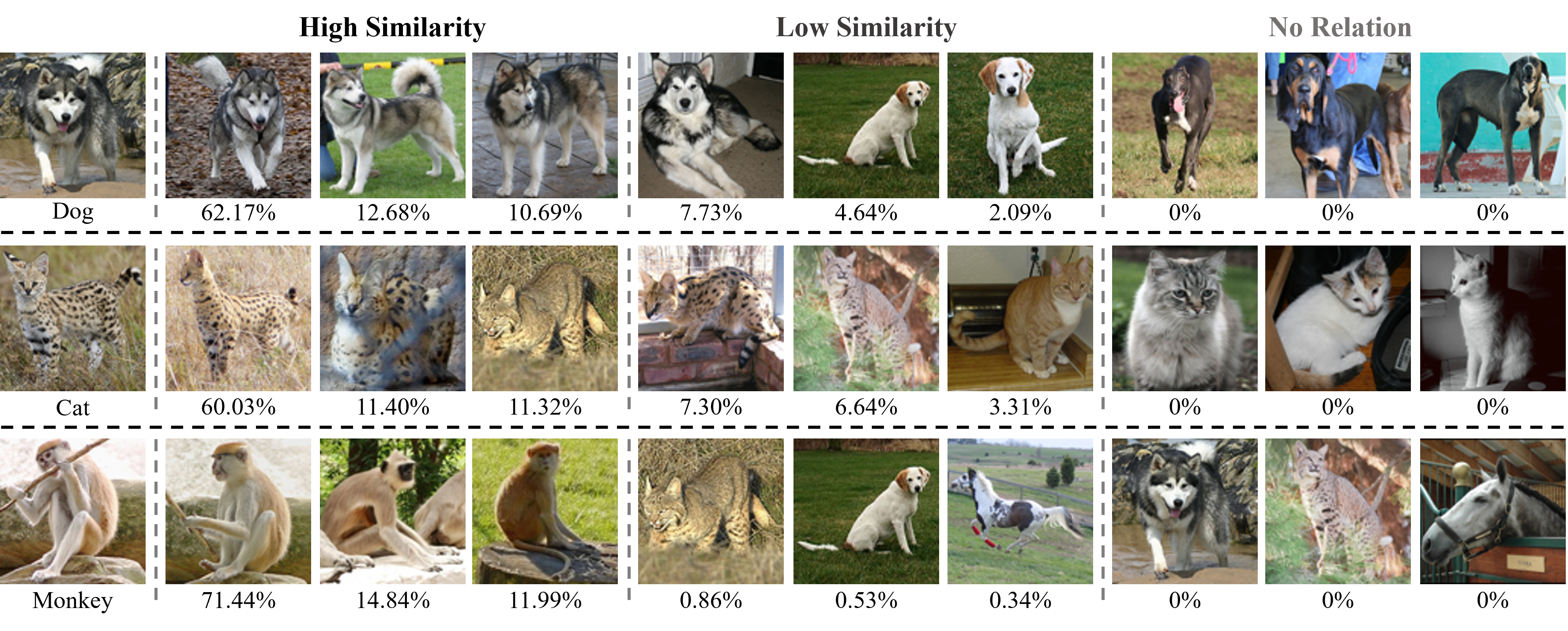}\\
	\vspace{-4mm}
	\caption{
		Visualization of the topological relationship learned by the differentiable sparse graph head on STL-10. The sparsity $s$ is $3$. The first and second rows show that the graph head can accurately learn the relationship among the instances in the same class. The bottom row suggests that the instance in different classes will be assigned a small similarity even $0$.
	}
	\vspace{-5mm}
	\label{sparsity}
\end{figure*} 

\section{Method} 
To bridge CNN and GNN, we propose a novel CNN2GNN heterogeneous distillation framework. It designs a deep graph learning head to differentiable learn the graph with the steerable sparsity according to the downstream tasks and eliminate the obstacle between CNN and GNN. Then, by the response-based distillation, the distilled ``boosted'' GNN will inductively extract the intra-sample representation of a single instance and the topological relationship among the instances simultaneously. The framework is shown in Fig \ref{framework}.

\subsection{Motivation}
With the assistance of two crucial components, the full connectivity layer and convolutional operator, \textbf{convolution neural networks} (CNNs) can project the image instance into the deep subspace to learn the intra-sample representation and have achieved great performance in image-related tasks. Among them, the full connectivity layer can be formulated as the linear model as
\begin{equation}
	\label{CNN}
	{\rm Linear}(\bm x)=\bm w^T \bm x,
\end{equation}
where $\bm w$ is a learning parameter. Meanwhile, the convolutional operator shares the weights across the channels \cite{goodfellow2016deep}, so it can be also reformulated as a linear transformation like im2col \cite{chellapilla2006high}. Thus, CNNs generally stack dozens of these operations and contain plenty of parameters for optimization to improve the performance, which may takes vast resources for optimization and storage. Besides, it is the linear projection that would limit the CNNs to explore the different information such as the relationship (similarity or differences) among the samples. 

Recently, \textbf{graph neural networks} (GNNs) have shown the excellent representational ability to learn the underlying information among the instances thanks to the graph dependency. Given a graph $\bm A$, a GNN layer is 
\begin{equation}
	\label{GNN}
	\bm Z = f_g(\bm A, \bm X, \bm W)=\varphi(\bm P \bm X \bm W),
\end{equation}
where $\bm W$ is the projection parameter and $\bm P=\phi(\bm A)$ is a function of $\bm A$. Similar to \cite{kipf2016semi}, $\phi(\bm A)={\bm D}^{-\frac{1}{2}} {\bm A} {\bm D}^{-\frac{1}{2}}$ and ${\bm D}$ is the degree matrix of $\bm A$. Notably, if the activation function $\varphi(\cdot)$ is ReLU, the deep feature generated from Eq. (\ref{GNN}) can be linear with multiple factors including a graph factor $\bm P$ and a projection factor $\bm W$. Among them, $\bm W$ will work like $\bm w$ in CNNs to learn the deep information of the single instance and $\bm P=\phi(\bm A)$ will aggregate the information from the neighbors to extract the latent structure information among the samples. That is shown that GNN as a bilinear model can represent more information with few neural layers compared with the linear transformation in CNN such as the full connectivity and im2col.  

Motivated by the bilinear property of GNNs that introduce more linear factors and represent more information in feature space compared with CNNs, we rethink the substantial difference between CNNs and GNNs. To sum up, the core question raised in this paper is \textbf{\textit{how to bridge these two heterogeneous networks, CNN and GNN?}}

Unfortunately, there are two critical issues in handling this problem, (1) graph inaccessible and (2) graph dependency. Specifically, since the graph factor is not prior information for non-graph data, GNN may not be directly applied to the image-related tasks. Although the vector product or Euclidean distance can construct a graph \cite{caramalau2021sequential, mohamed2020social}, it will take much time to select a proper sparse threshold and the obtained graph lacks information about the downstream tasks due to independence with the gradient descent. Besides, with the limitation of the graph dependency, GNN generally is trained with transductive learning and has a high inference latency such as $O(R^l)$ \cite{zhang2021graph}, where $R$ is the graph degree and $l$ is the number of layers. It is impractical to extend on the large scenario where $R$ is large. In this paper, we will firstly handle these two critical issues in Section \ref{sparse_graph} and elaborate on the core question in Section \ref{distillation}.

\subsection{Differentiable Sparse Graph Head} \label{sparse_graph}
To bridge these two heterogeneous networks, a natural thought is substituting the CNN with the GNN model and improving the representational ability by introducing the two linear factors in a single neural layer. Unfortunately, since the graph structure is not directly accessible for non-graph datasets such as images, GNN cannot be extended on the image data directly. Besides, due to aggregating the information recursively from the sample neighbors, GNN needs to be trained with transductive learning and the inference latency grows exponentially with the graph degree such as $O(R^l)$, where $R$ is the graph degree and $l$ is the number of layers. It causes the transductive learning strategy will be impractical when $R$ is large in real-world scenarios. Therefore, we firstly design a differentiable sparse graph head to \textit{inductively learn the graph from the non-graph data}.

\textbf{Sparse Graph Learning:} Give a non-graph dataset $\mathcal{X}$ with $n$ samples and a single sample $\bm x$ is viewed as one node $\bm v$ in a graph. Besides, we regard the underlying graph structure of a single sample $\bm v_i$ as the conditional probability $p(\bm v | \bm v_i)$ and the edge as a sampling result from this distribution $p(\bm v | \bm v_i)$. Therefore, the generation of graph is equivalent to calculating the distribution $p(\bm v | \bm v_i)$ as
\begin{equation}
	\label{graph_solve}
	p_{ij}=(\frac{\left\langle \bm v_i, \bm v^{s+1} \right\rangle-\left\langle \bm v_i, \bm v_j \right\rangle}{\sum_{j=1}^s \left\langle \bm v_i, \bm v^{s+1} \right\rangle-\left\langle \bm v_i, \bm v_j \right\rangle})_+, \
\end{equation}
where $s$ is the sparsity, $(\cdot)_+$ means $\max(\cdot, 0)$, and $p(\cdot | \bm v_i)$ is simplified as $\bm p_i$. In Eq. (\ref{graph_solve}), $\left\langle \bm v_i, \bm v_j \right\rangle$ is the distance from $\bm v_j$ to $\bm v_i$ and $\left\langle \bm v_i, \bm v^{\cdot} \right\rangle$ represents the $\cdot$-th smallest distance to $\bm v_i$. Then, the undirected sparse graph is constructed as $\bm A_{ij}=(p(\bm v_j | \bm v_i) + p(\bm v_i | \bm v_j))/2$. Compared with utilizing the vector inner product ($\bm v_i^T \bm v_j$) \cite{caramalau2021sequential} or the Euclidean distance ($\|\bm v_i - \bm v_j\|_2$) \cite{mohamed2020social} to generate the graph structure and control the sparsity by selecting extra hyper-parameters, Eq. (\ref{graph_solve}) can explicitly control the learned graph sparsity and save the tuning cost. Theorem \ref{transform_pro} indicates that Eq. (\ref{graph_solve}) is the sparse closed-form solution of $p(\bm v | \bm v_i)$. Notably, $\ell_2$-norm is employed as the regularization term in Eq. (\ref{construct_l2}). Compared with the $\ell_0$-norm non-convex constriant and the relaxation $\ell_1$-norm regularization, it not only prevents the trivial solution such as $p(\bm v_i | \bm v_i)=1$ and $p(\bm v_i | \bm v_j)=0$ if $i \neq j$, but also gaurantees steerable sparsity. The proof of Theorem \ref{transform_pro} is in the supplementary.

Furthermore, considering that an ideal graph could be dynamically and differentiable learned during the optimization to contain more the downstream task information, we utilize a deep neural network as the graph generation head $f_{\star}^g(\cdot)$ to fit $\left\langle \bm v_i, \bm v_j \right\rangle$ and will be jointly optimized with GNN. It has $n$ neurons in the decision layer and each neuron represents the difference with the corresponding sample in $\mathcal{X}$, where $n$ is the total number of the samples. Based on this, the distance is fitted with $\left\langle \bm v_i, \mathcal{X} \right\rangle=f_{\star}^g(\bm x_i)$ and the sparse graph will be calculated via Eq. (\ref{graph_solve}). Meanmwhile, as shown in Fig. \ref{sparsity}, Eq. (\ref{graph_solve}) can accurately learn the relationship in STL-10.

\textbf{Inductive Learning:} Notably, it is the graph structure obtained dynamically that GNN can be trained with inductive learning. Split the dataset $\mathcal{X}$ into a training indices set $\mathcal{I}_{\rm train}$ with $n_{\rm train}$ instances and the testing indices set $\mathcal{I}_{\rm test}$ with $n_{\rm test}$ instances. And the sparse graph generation $f_{\star}^g(\cdot)$ is equipped with $n_{\rm train}$ neurons. 

\textbf{1) Training Stage:} For a training batch $\mathcal{B} \subseteq \mathcal{I}_{\rm train}$, the $s$-sparse similarity distribution is
\begin{equation}
	\label{batch_graph}
	p_{ij}=(\frac{\left\langle \bm v_i, \bm v^{s+1} \right\rangle-\left\langle \bm v_i, \bm v_j \right\rangle}{\sum_{j=1}^s \left\langle \bm v_i, \bm v^{s+1} \right\rangle-\left\langle \bm v_i, \bm v_j \right\rangle})_+, i \in \mathcal{B}, j \in \mathcal{I}_{\rm train},
\end{equation}
where $\left\langle \bm v_i, \bm v \right\rangle$ is fitted by $f_{\star}^g(\bm v_i)$. Thus, we can obtain the graph structure via $\bm A_{ij}=(p_{ij} + p_{ji})/2$ and calculate the representation of $\mathcal{B}$ via Eq. (\ref{GNN}). Notably, since all operations are differentiable, they can be optimized with gradient descent. 

\textbf{2) Testing Stage:} Since $f_{\star}^g(\cdot)$ cannot measure the distance between $\bm v_i$ and $\bm v_j$ where $i, j \in \mathcal{I}_{\rm test}$, we introduce a training batch $\mathcal{B}_{\rm train}$ and design two mechanisms to induce and learn the testing similarity distribution. Among them, the first one cascades a test instance with a training batch and learns the graph to evaluate the performance of the testing set, which is shown in Mechanism \ref{one_instance}. However, it only tests one instance in each iteration, which is inefficient on large datasets. Therefore, to improve the efficiency, Mechanism \ref{batch_instance} selects the most similar sample in the training set to learn an approximated graph structure for evaluating the testing samples batch-by-batch. 
\begin{myMech}
	\label{one_instance}
	Give a testing sample $\bm v_i,~i \in \mathcal{I}_{\rm test}$, it firstly is cascaded with a set of training samples to form batch $\mathcal{B}$. Then, the sparse graph is learned as $\bm A_{ij}=(p_{ij} + p_{ji})/2$, where $p_{ij}$ is calculated by Eq. (\ref{graph_solve}). Thus, GNN can inference the representation of $\bm v_i$.
\end{myMech}
\begin{myMech}
	\label{batch_instance}
	Given a testing batch $\mathcal{B} \subseteq \mathcal{I}_{\rm test}$, the conditional distribution $\bm P$ between the testing batch $\mathcal{B}_{\rm test}$ and the training batch $\mathcal{B}_{\rm train}$ is calculated via Eq. (\ref{graph_solve}). Then,  the approximated batch $\mathcal{B}_{\rm sim}$ is formed by choosing the highest probability in $\bm P$. Finally, the approximated sparse graph is learned as $\bm A_{ij}^{\rm{apr}}=(p_{ij}^{\rm{apr}} + p_{ji}^{\rm{apr}})/2$, where $p_{ij}^{\rm{apr}}$ is calculated by Eq. (\ref{graph_solve}). $\bm A_{ij}^{\rm{apr}}$ assists GNN to generate the representation of the testing batch.
\end{myMech}

In a word, Mechanism \ref{one_instance} can directly obtain the graph structure of the testing sample $\bm v_i$ but it has the lower inference efficiency when $n_{\rm test}$ is large. Meanwhile, although Mechanism \ref{batch_instance} can improve the efficiency of the testing set, the approximated strategy may sacrifice performance. Furthermore, we will detailly discuss the merits and limitations in Section \ref{ablation}. The whole procedure of the proposed graph learning head is suggested in Algorithm \ref{Graph Learning Head}. 

\begin{algorithm}[t]             
	\caption{Differentiable Sparse Graph Head for Inductive Learning}
	\label{Graph Learning Head}
	\begin{algorithmic}[1] 
		\REQUIRE Non-graph data $\mathcal{X}$, sparsity $s$, objective function $\mathcal{L}$ and GNN $f_{\star}(\cdot)$.\\
		\STATE Split the dataset $\mathcal{X}$ into a training indices set $\mathcal{I}_{\rm train}$ and the testing indices set $\mathcal{I}_{\rm test}$;
		\STATE Initialize GNN $f_{\star}(\cdot)$ and graph head $f_{\star}^g(\cdot)$ randomly;
		\WHILE{$not \; Convergence$}
		\STATE Obtain the graph $\bm A$ via Eq. (\ref{graph_solve}) on batch $\mathcal{B} \subseteq \mathcal{I}_{\rm train}$;
		\STATE Generate the graph representation $f_{\star}(\bm v_i; \bm A),~i\in \mathcal{B}$;
		\STATE Optimize $\mathcal{L}$ via gradient descent;
		\ENDWHILE
		\STATE Inference on the testing set $\mathcal{I}_{\rm test}$ via Mechanism \ref{one_instance} or  \ref{batch_instance};
	\end{algorithmic}
\end{algorithm}

%
%
%
%
%
%
%
%

\begin{myTheo}
	\label{transform_pro}
	Given a set of samples $\mathcal{V}=\left\{ \bm v_i | n=1,...,n \right\}$, the conditional probability $p(\bm v | \bm v_i)$ can be formulated from 
	\begin{equation}
		\label{construct_l2}
		\begin{split}
			\min_{\bm p_i^T \bm 1_n = 1,\bm p_i \geq 0} &\sum_{i=1}^{n} \mathbb{E}_{\bm v_j \sim p(\cdot | \bm v_i)} \left\langle \bm v_i, \bm v_j \right\rangle + \gamma_i {\rm dist}(\bm p_i, \bm \pi), \\
		\end{split}
	\end{equation}
	where $\bm \pi$ is a uniform distribution, ${\rm dist}(\cdot, \cdot)$ represents the $\ell_2$-norm distance, and $\gamma_i$ is the trade-off parameter. And Eq. (\ref{graph_solve}) is equivalent to a solution form of this problem.
\end{myTheo}

\subsection{CNN2GNN: Heterogeneous Distillation} \label{distillation}
After eliminating the obstacle between CNN and GNN by the proposed graph head, we introduce a response-based heterogeneous distillation to transfer the knowledge and bridge these two heterogeneous neural networks.

Specifically, we employ CNN $f_*(\cdot)$ and GNN $f_{\star}(\cdot)$ as the teacher and student, respectively. Among them, $f_*(\cdot)$ is a large network with plenty of layers such as ResNet-152 and GNN $f_{\star}(\cdot)$ only has two layers. Meanwhile, a instance $\bm x$ is viewed as one node $\bm v$ in a graph. Firstly, the CNN teacher is trained by cross-entropy loss to extract the deep intra-sample representation $f_*(\bm x)$. Then, for a training batch $\mathcal{B}\subseteq \mathcal{I}_{\rm train}$, the differentiable sparse graph head provides the graph $\bm A$ for a GNN student to learn the relationship representation $f_{\star}(\bm v; \bm A)$ among the instance sets. The objective function is defined as
\begin{equation}
	\label{student_batch}
	\begin{split}
		\mathcal{L}_{\rm student}&=\frac{1}{|\mathcal{B}|}\sum_{i \in \mathcal{B}} \mathcal{L}_{\rm cross}(f_{\star}(\bm v_i; \bm A), \bm y_i)\\
		&+\alpha \mathcal{L}_{\rm kd}(f_{\star}(\bm v_i; \bm A), f_*(\bm x_i), \tau),  \mathcal{B} \subseteq \mathcal{I}_{\rm train},\\
	\end{split}
\end{equation}
where $\bm y_i$ is the label, $\alpha$ is a weight parameter, and $\tau$ is a temperature. $\mathcal{L}_{\rm cross}(\cdot)$ is the cross-entropy loss, which mainly guides the GNN to explore the latent relationship information and generate the reliable graph representation. $\mathcal{L}_{\rm kd}$ is a response-based distillation \cite{hinton2015distilling} as
\begin{equation} 
	\label{kd_loss}
	\mathcal{L}_{\rm kd}= {\rm{KL}}(q(f_*(\bm v_i),\tau) \| q(f_{\star}(\bm v_i; \bm A),\tau) ),
\end{equation}
where  $q(\cdot_v,\tau)=\frac{{\rm{exp}}({\cdot_v}/{\tau})}{\sum_j{\rm{exp}}({\cdot_j}/{\tau})}$, $f_{\star}(\bm x_v) = \bm h_v^l$ and ${\rm{KL}}(\cdot \| \cdot)$ is a KL-divergence. 
Meanwhile, by introducing $f_*(\bm v_i)$ as the soft target, the deep intra-sample representation can be distilled into the GNN and these two heterogeneous neural networks are bridged in the temperature $\tau$. After training, with the assistance of Mechanism \ref{one_instance} or Mechanism \ref{batch_instance}, the distilled ``boosted'' GNN can inductively extract the local representation of a single instance and the topological relationship among the instance sets simultaneously. The whole optimization of CNN2GNN is shown in Algorithm \ref{CNN2GNN distillation}.

\begin{algorithm}[t]             
	\caption{Heterogeneous Distillation: Bridge CNN and GNN}
	\label{CNN2GNN distillation}
	\begin{algorithmic}[1] 
		\REQUIRE Non-graph data $\mathcal{X}$, the groundtruth $\mathcal{Y}$, sparsity $s$, trade-off parameter $\alpha$, temperature $\tau$, CNN $f_{*}(\cdot)$ and GNN $f_{\star}(\cdot)$.\\
		\STATE Split the dataset $\mathcal{X}$ into a training indices set $\mathcal{I}_{\rm train}$ and the testing indices set $\mathcal{I}_{\rm test}$;
		\STATE Initialize $f_{*}(\cdot)$, $f_{\star}(\cdot)$ and graph head $f_{\star}^g(\cdot)$ randomly;
		\STATE Pretrian CNN teacher $f_{*}(\cdot)$ via gradient descent;
		\WHILE{$not \; Convergence$}
		\STATE Obtain the graph $\bm A$ via Eq. (\ref{graph_solve}) on batch $\mathcal{B} \subseteq \mathcal{I}_{\rm train}$;
		\STATE GNN student generates the graph representation $f_{\star}(\bm v_i; \bm A),~i\in \mathcal{B}$;
		\STATE CNN teacher generates the deep intra-sample representation $f_{*}(\bm x_i),~i\in \mathcal{B}$;
		\STATE Optimize $\mathcal{L}_{\rm student}$ via gradient descent;
		\ENDWHILE
		\STATE Distilled ``boosted'' GNN inductively inference on $\mathcal{I}_{\rm test}$ via Mechanism \ref{one_instance} or Mechanism \ref{batch_instance};
	\end{algorithmic}
\end{algorithm}


\noindent \textbf{The Merits of the Proposed Model: } 
Inspired by the complementary merits of two mainstream neural networks, we distill a large CNN into a tiny GNN and bridge these two heterogeneous networks. Firstly, a graph head is designed to differentiable and inductively learn a sparse graph structure, which can extend the GNN on non-graph data and eliminate the obstacles between these two networks. Then, with the response-based heterogeneous distillation, the distilled ``boosted'' GNN with a few neural layers can inductively extract the deep intra-sample representation of a single instance and the topological relationship among the instance sets simultaneously.

\begin{table*}[t]
	\renewcommand\arraystretch{1.2}
	\centering
	\caption{Accuracy(\%) on CIFAR-100}
	\vspace{-3mm}
	\label{Results_cifar100}
	\scalebox{0.8}{
		\begin{tabular}{cccccccccc}
			\toprule
			\textbf{\begin{tabular}[c]{@{}c@{}}Teacher\\ Student\end{tabular}} & \textbf{\begin{tabular}[c]{@{}c@{}}ResNet34\\ ResNet18\end{tabular}} & \textbf{\begin{tabular}[c]{@{}c@{}}ResNet50\\ ResNet18\end{tabular}} & \textbf{\begin{tabular}[c]{@{}c@{}}ResNet101\\ ResNet18\end{tabular}} & \textbf{\begin{tabular}[c]{@{}c@{}}ResNet152\\ ResNet18\end{tabular}} & \textbf{\begin{tabular}[c]{@{}c@{}}VGG13\\ VGG8\end{tabular}} & \textbf{\begin{tabular}[c]{@{}c@{}}VGG16\\ VGG8\end{tabular}} & \textbf{\begin{tabular}[c]{@{}c@{}}VGG19\\ VGG8\end{tabular}} & \textbf{\begin{tabular}[c]{@{}c@{}}WRN-40-2\\ WRN-16-2\end{tabular}} & \textbf{\begin{tabular}[c]{@{}c@{}}WRN-40-2\\ WRN-40-1\end{tabular}} \\ \hline
			Teacher & 62.19	& 63.91 & 65.19 & 66.00 & 58.48 & 59.25 & 59.16 & 55.32 & 55.32 \\ 
			KD & 62.78	& 62.25 & 62.32 & 62.61 & 57.15 & 57.97 & 57.72 & 54.86 & 55.03 \\ \hline
			FitNets & 62.38 ({\color{red}$\downarrow$})	& 60.73 ({\color{red}$\downarrow$}) & 62.34 ({\color{green}$\uparrow$}) & 61.89 ({\color{red}$\downarrow$}) & 57.06 ({\color{red}$\downarrow$}) & 57.90 ({\color{red}$\downarrow$}) & 57.63 ({\color{red}$\downarrow$}) & 55.04 ({\color{green}$\uparrow$}) & 55.93 ({\color{green}$\uparrow$}) \\ 
			AT & 62.19 ({\color{red}$\downarrow$})	& 60.46 ({\color{red}$\downarrow$}) & 62.61 ({\color{green}$\uparrow$}) & 62.28 ({\color{red}$\downarrow$}) & 57.10 ({\color{red}$\downarrow$}) & 57.73 ({\color{red}$\downarrow$}) & 58.02 ({\color{green}$\uparrow$}) & 54.04 ({\color{green}$\uparrow$}) & 55.35 ({\color{green}$\uparrow$}) \\ 
			SP & 62.24 ({\color{red}$\downarrow$})	& 61.94 ({\color{red}$\downarrow$}) & 61.94 ({\color{red}$\downarrow$}) & 62.82 ({\color{green}$\uparrow$}) & 56.79 ({\color{red}$\downarrow$}) & 57.82 ({\color{red}$\downarrow$}) & 56.84 ({\color{red}$\downarrow$}) & 54.14 ({\color{red}$\downarrow$}) & 54.61 ({\color{red}$\downarrow$}) \\
			VID & 62.96 ({\color{green}$\uparrow$})	& 61.58 ({\color{red}$\downarrow$}) & 62.44 ({\color{green}$\uparrow$}) & 63.03 ({\color{green}$\uparrow$}) & 57.15 ({\color{green}$\uparrow$}) & 58.21 ({\color{green}$\uparrow$}) & 58.32 ({\color{green}$\uparrow$}) & 57.02 ({\color{green}$\uparrow$}) & 56.06 ({\color{green}$\uparrow$}) \\
			RKD & 63.01 ({\color{green}$\uparrow$})	& 63.58 ({\color{green}$\uparrow$}) & 63.53 ({\color{green}$\uparrow$}) & 64.03 ({\color{green}$\uparrow$}) & 57.34 ({\color{green}$\uparrow$}) & 58.08 ({\color{green}$\uparrow$}) & \textbf{59.13} ({\color{green}$\uparrow$}) & 56.06 ({\color{green}$\uparrow$}) & 57.46 ({\color{green}$\uparrow$}) \\
			PKT & 62.65 ({\color{red}$\downarrow$})	& 61.89 ({\color{red}$\downarrow$}) & 62.13 ({\color{red}$\downarrow$}) & 62.17 ({\color{red}$\downarrow$}) & 57.37 ({\color{green}$\uparrow$}) & 56.09 ({\color{red}$\downarrow$}) & 56.66 ({\color{red}$\downarrow$}) & 55.14 ({\color{green}$\uparrow$}) & 55.27 ({\color{green}$\uparrow$}) \\
			AB & 57.23 ({\color{red}$\downarrow$})	& 62.14 ({\color{red}$\downarrow$}) & 63.14 ({\color{red}$\downarrow$}) & 62.00 ({\color{red}$\downarrow$}) & 56.75 ({\color{red}$\downarrow$}) & 57.50 ({\color{red}$\downarrow$}) & 56.47 ({\color{red}$\downarrow$}) & 53.58 ({\color{red}$\downarrow$}) & 55.01 ({\color{red}$\downarrow$}) \\
			FT & 62.04 ({\color{red}$\downarrow$})	& 63.04 ({\color{green}$\uparrow$}) & 62.31 ({\color{red}$\downarrow$}) & 62.41 ({\color{red}$\downarrow$}) & 57.57 ({\color{green}$\uparrow$}) & 57.86 ({\color{red}$\downarrow$}) & 57.68 ({\color{red}$\downarrow$}) & 54.38 ({\color{red}$\downarrow$}) & 54.95 ({\color{red}$\downarrow$}) \\
			CRD & 57.95 ({\color{red}$\downarrow$})	& 58.00 ({\color{red}$\downarrow$}) & 57.56 ({\color{red}$\downarrow$}) & 57.57 ({\color{red}$\downarrow$}) & 51.03 ({\color{red}$\downarrow$}) & 51.95 ({\color{red}$\downarrow$}) & 51.79 ({\color{red}$\downarrow$}) & 51.30 ({\color{red}$\downarrow$}) & 52.84 ({\color{red}$\downarrow$})  \\
			CCD & 63.17 ({\color{green}$\uparrow$})	& 64.06 ({\color{green}$\uparrow$}) & \underline{64.41} ({\color{green}$\uparrow$}) & 64.03 ({\color{green}$\uparrow$}) & 58.95 ({\color{green}$\uparrow$}) & \underline{58.50} ({\color{green}$\uparrow$}) & 58.32 ({\color{green}$\uparrow$}) & \underline{62.26} ({\color{green}$\uparrow$}) & \underline{65.05} ({\color{green}$\uparrow$}) \\
			DKD & \underline{64.15} ({\color{green}$\uparrow$})	& \underline{65.15} ({\color{green}$\uparrow$}) & 63.72 ({\color{green}$\uparrow$}) & \underline{64.43} ({\color{green}$\uparrow$}) & \underline{58.97} ({\color{green}$\uparrow$}) & 57.78 ({\color{red}$\downarrow$}) & 57.19 ({\color{red}$\downarrow$}) & 61.16 ({\color{green}$\uparrow$}) & 64.30 ({\color{green}$\uparrow$}) \\ \hline
			\textbf{GNN} & 60.21 ({\color{red}$\downarrow$})	& 61.01 ({\color{red}$\downarrow$}) & 60.28 ({\color{red}$\downarrow$}) & 60.06 ({\color{red}$\downarrow$}) & 56.40 ({\color{red}$\downarrow$}) & 55.84 ({\color{red}$\downarrow$}) & 56.36 ({\color{red}$\downarrow$}) & 60.76 ({\color{green}$\uparrow$}) & 60.76 ({\color{green}$\uparrow$}) \\
			\textbf{CNN2GNN} & \textbf{65.13} ({\color{green}$\uparrow$})	& \textbf{65.53} ({\color{green}$\uparrow$}) & \textbf{64.43} ({\color{green}$\uparrow$}) & \textbf{64.67} ({\color{green}$\uparrow$}) & \textbf{59.33} ({\color{green}$\uparrow$}) & \textbf{58.68} ({\color{green}$\uparrow$}) & \underline{58.57} ({\color{green}$\uparrow$}) & \textbf{65.58} ({\color{green}$\uparrow$}) & \textbf{65.36} ({\color{green}$\uparrow$}) \\ 
			\bottomrule
		\end{tabular}
	}
	\vspace{-5mm}
\end{table*}

\begin{figure*}[t]
	\centering
	\subfigure[ResNet50-18]{
		\label{ResNet50-18-Cifar100}
		\includegraphics[scale=0.23]{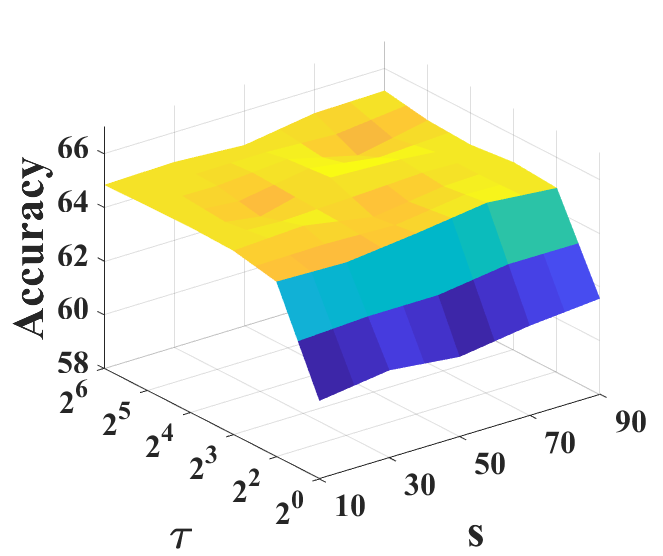}
	}
	\subfigure[VGG13-8]{
		\label{VGG13-8-Cifar100}
		\includegraphics[scale=0.23]{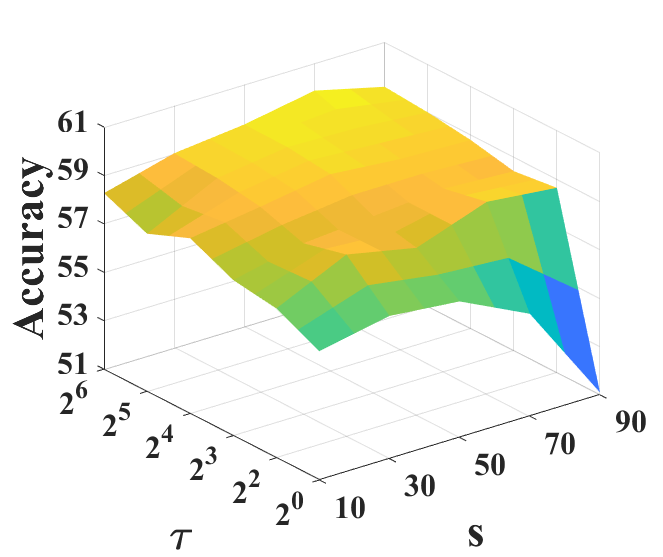}
	}
	\subfigure[ResNet50-18]{
		\label{ResNet50-18-ImageNet}
		\includegraphics[scale=0.23]{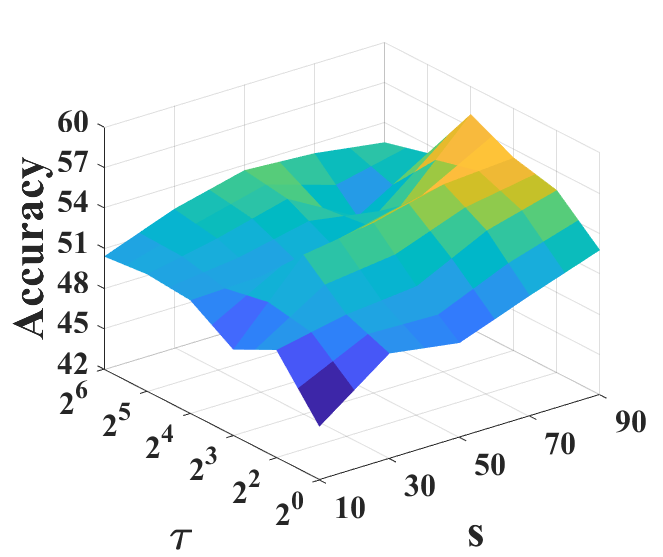}
	}
	\subfigure[VGG13-8]{
		\label{VGG13-8-ImageNet}
		\includegraphics[scale=0.23]{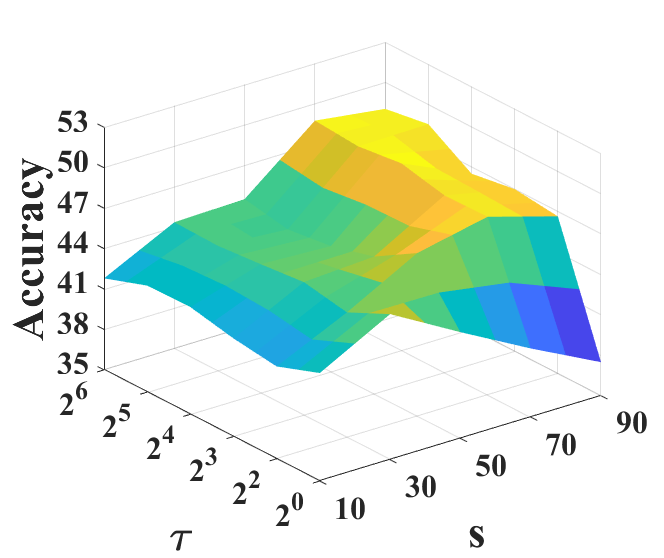}
	}
	\vspace{-3mm}
	\caption{Accuracy of CNN2GNN w.r.t the varying parameter $\tau \in \left\{ 2^{0}, 2^{2}, 2^{3}, 2^4, 2^5, 2^6 \right\}$ and $s \in \left\{ 10, 30, 50, 70, 90 \right\}$. Figure \ref{ResNet50-18-Cifar100} and \ref{VGG13-8-Cifar100} are the results on CIFAR-100. Figure \ref{ResNet50-18-ImageNet} and \ref{VGG13-8-ImageNet} are the results on Mini-Imagenet.}
	\label{Sensitivity}
	\vspace{-5mm}	
\end{figure*}

\begin{table*}[t]
	\renewcommand\arraystretch{1.2}
	\centering
	\caption{Accuracy(\%) on Mini-ImageNet}
	\vspace{-3mm}
	\label{Results_miniimagenet}
	\scalebox{0.8}{
		\begin{tabular}{cccccccccc}
			\toprule
			\textbf{\begin{tabular}[c]{@{}c@{}}Teacher\\ Student\end{tabular}} & \textbf{\begin{tabular}[c]{@{}c@{}}ResNet34\\ ResNet18\end{tabular}} & \textbf{\begin{tabular}[c]{@{}c@{}}ResNet50\\ ResNet18\end{tabular}} & \textbf{\begin{tabular}[c]{@{}c@{}}ResNet101\\ ResNet18\end{tabular}} & \textbf{\begin{tabular}[c]{@{}c@{}}ResNet152\\ ResNet18\end{tabular}} & \textbf{\begin{tabular}[c]{@{}c@{}}VGG13\\ VGG8\end{tabular}} & \textbf{\begin{tabular}[c]{@{}c@{}}VGG16\\ VGG8\end{tabular}} & \textbf{\begin{tabular}[c]{@{}c@{}}VGG19\\ VGG8\end{tabular}} & \textbf{\begin{tabular}[c]{@{}c@{}}WRN-40-2\\ WRN-16-2\end{tabular}} & \textbf{\begin{tabular}[c]{@{}c@{}}WRN-40-2\\ WRN-40-1\end{tabular}} \\ \hline
			Teacher & 45.11	& 49.70 & 48.25 & 50.35 & 46.50 & 46.30 & 45.34 & 51.75 & 51.75 \\ 
			KD & 45.07	& 45.39 & 46.18 & 41.72 & 43.46 & 42.05 & 41.65 & 41.52 & 42.43 \\ \hline
			FitNets & 45.69 ({\color{green}$\uparrow$})	& 46.11 ({\color{green}$\uparrow$}) & 45.76 ({\color{red}$\downarrow$}) & 45.35 ({\color{green}$\uparrow$}) & 43.54 ({\color{green}$\uparrow$}) & 41.89 ({\color{red}$\downarrow$}) & 40.96 ({\color{red}$\downarrow$}) & 42.79 ({\color{green}$\uparrow$}) & 43.96 ({\color{green}$\uparrow$}) \\ 
			AT & 45.86 ({\color{green}$\uparrow$})	& 45.54 ({\color{green}$\uparrow$}) & 45.18 ({\color{red}$\downarrow$}) & 46.88 ({\color{green}$\uparrow$}) & 43.27 ({\color{red}$\downarrow$}) & 42.07 ({\color{green}$\uparrow$}) & 41.92 ({\color{green}$\uparrow$}) & 43.02 ({\color{green}$\uparrow$}) & 42.63 ({\color{green}$\uparrow$}) \\ 
			SP & 45.55 ({\color{green}$\uparrow$})	& 46.39 ({\color{green}$\uparrow$}) & 45.81 ({\color{red}$\downarrow$}) & 45.95 ({\color{green}$\uparrow$}) & 42.68 ({\color{red}$\downarrow$}) & 42.30 ({\color{green}$\uparrow$}) & 41.91 ({\color{green}$\uparrow$}) & 41.97 ({\color{green}$\uparrow$}) & 42.05 ({\color{red}$\downarrow$}) \\
			VID & 47.24 ({\color{green}$\uparrow$})	& 45.23 ({\color{red}$\downarrow$}) & 45.87 ({\color{red}$\downarrow$}) & 45.77 ({\color{green}$\uparrow$}) & 42.81 ({\color{red}$\downarrow$}) & 42.20 ({\color{green}$\uparrow$}) & 42.31 ({\color{green}$\uparrow$}) & 42.82 ({\color{green}$\uparrow$}) & 43.46 ({\color{green}$\uparrow$}) \\
			RKD & 47.23 ({\color{green}$\uparrow$})	& 49.30 ({\color{green}$\uparrow$}) & 49.12 ({\color{green}$\uparrow$}) & \underline{49.63} ({\color{green}$\uparrow$}) & 43.55 ({\color{green}$\uparrow$}) & 42.92 ({\color{green}$\uparrow$}) & 41.35 ({\color{red}$\downarrow$}) & 43.08 ({\color{green}$\uparrow$}) & 44.02 ({\color{green}$\uparrow$}) \\
			PKT & 45.66 ({\color{green}$\uparrow$})	& 46.09 ({\color{green}$\uparrow$}) & 46.05 ({\color{red}$\downarrow$}) & 45.84 ({\color{green}$\uparrow$}) & 42.65 ({\color{red}$\downarrow$}) & 41.81 ({\color{red}$\downarrow$}) & 41.21 ({\color{red}$\downarrow$}) & 42.04 ({\color{green}$\uparrow$}) & 42.30 ({\color{red}$\downarrow$}) \\
			AB & 41.37 ({\color{red}$\downarrow$})	& 46.14 ({\color{green}$\uparrow$}) & 45.22 ({\color{red}$\downarrow$}) & 45.68 ({\color{green}$\uparrow$}) & 43.66 ({\color{green}$\uparrow$}) & 40.55 ({\color{red}$\downarrow$}) & 38.31 ({\color{red}$\downarrow$}) & 43.10 ({\color{green}$\uparrow$}) & 43.71 ({\color{green}$\uparrow$}) \\
			FT & 46.50 ({\color{green}$\uparrow$})	& 45.01 ({\color{red}$\downarrow$}) & 45.75 ({\color{red}$\downarrow$}) & 46.33 ({\color{green}$\uparrow$}) & 42.74 ({\color{red}$\downarrow$}) & 41.57 ({\color{red}$\downarrow$}) & 41.36 ({\color{red}$\downarrow$}) & 42.93 ({\color{green}$\uparrow$}) & 42.77 ({\color{green}$\uparrow$}) \\
			CRD & 46.03 ({\color{green}$\uparrow$})	& 41.73 ({\color{red}$\downarrow$}) & 38.91 ({\color{red}$\downarrow$}) & 41.75 ({\color{green}$\uparrow$}) & 41.18 ({\color{red}$\downarrow$}) & 40.79 ({\color{red}$\downarrow$}) & 40.40 ({\color{red}$\downarrow$}) & 41.93 ({\color{green}$\uparrow$}) & 41.49 ({\color{red}$\downarrow$}) \\
			CCD & \underline{48.64} ({\color{green}$\uparrow$}) & \underline{49.04} ({\color{green}$\uparrow$}) & \underline{49.84} ({\color{green}$\uparrow$}) & 49.44 ({\color{green}$\uparrow$}) & 43.68 ({\color{green}$\uparrow$}) & 43.60 ({\color{green}$\uparrow$}) & 41.75 ({\color{green}$\uparrow$}) & 51.07 ({\color{green}$\uparrow$}) & \underline{51.79} ({\color{green}$\uparrow$}) \\
			DKD & 47.69 ({\color{green}$\uparrow$}) & {48.35} ({\color{green}$\uparrow$}) & 47.70 ({\color{green}$\uparrow$}) & 46.86 ({\color{green}$\uparrow$}) & \underline{45.50} ({\color{green}$\uparrow$}) & \underline{45.23} ({\color{green}$\uparrow$}) & \underline{44.27} ({\color{green}$\uparrow$}) & \underline{52.11} ({\color{green}$\uparrow$}) & 50.41 ({\color{green}$\uparrow$}) \\ \hline
			\textbf{GNN} & 44.27 ({\color{red}$\downarrow$}) & 43.84 ({\color{red}$\downarrow$}) & 43.87 ({\color{red}$\downarrow$}) & 44.02 ({\color{green}$\uparrow$}) & 38.89 ({\color{red}$\downarrow$}) & 38.40 ({\color{red}$\downarrow$}) & 38.27 ({\color{red}$\downarrow$}) & 42.36 ({\color{green}$\uparrow$}) & 42.36 ({\color{red}$\downarrow$}) \\
			\textbf{CNN2GNN} & \textbf{57.94} ({\color{green}$\uparrow$})	& \textbf{55.04} ({\color{green}$\uparrow$}) & \textbf{53.44} ({\color{green}$\uparrow$}) & \textbf{60.64} ({\color{green}$\uparrow$}) & \textbf{48.88} ({\color{green}$\uparrow$}) & \textbf{46.57} ({\color{green}$\uparrow$}) & \textbf{46.16} ({\color{green}$\uparrow$}) & \textbf{53.09} ({\color{green}$\uparrow$}) & \textbf{52.72} ({\color{green}$\uparrow$}) \\ 
			\bottomrule
		\end{tabular}
	}
	\vspace{-3mm}
\end{table*}

\section{Experiment}

\subsection{Experimental Settings} \label{dataset description}
\noindent \textbf{Dataset and Baseline:} We experiment on four real-world scenario image datasets including STL-10 \cite{coates2011analysis}, CIFAR-10 \cite{krizhevsky2009learning}, CIFAR-100, and Mini-ImageNet \cite{deng2009imagenet} with various teacher-student combinations such as VGG \cite{simonyan2014very}, ResNet \cite{he2016deep} and WRN \cite{zagoruyko2016wide}. KD \cite{hinton2015distilling}, FitNets \cite{romero2014fitnets}, AT \cite{komodakis2017paying}, SP \cite{tung2019similarity}, VID \cite{ahn2019variational}, RKD \cite{park2019relational}, PKT \cite{passalis2018learning}, AB \cite{heo2019knowledge}, FT \cite{kim2018paraphrasing}, CRD \cite{tian2019contrastive}, CCD \cite{li2022role} and DKD \cite{zhao2022decoupled} are introduced as the baselines. The detailed data pre-processing is in the supplementary.

\begin{figure}[t]
	\centering
	\includegraphics[width=70mm]{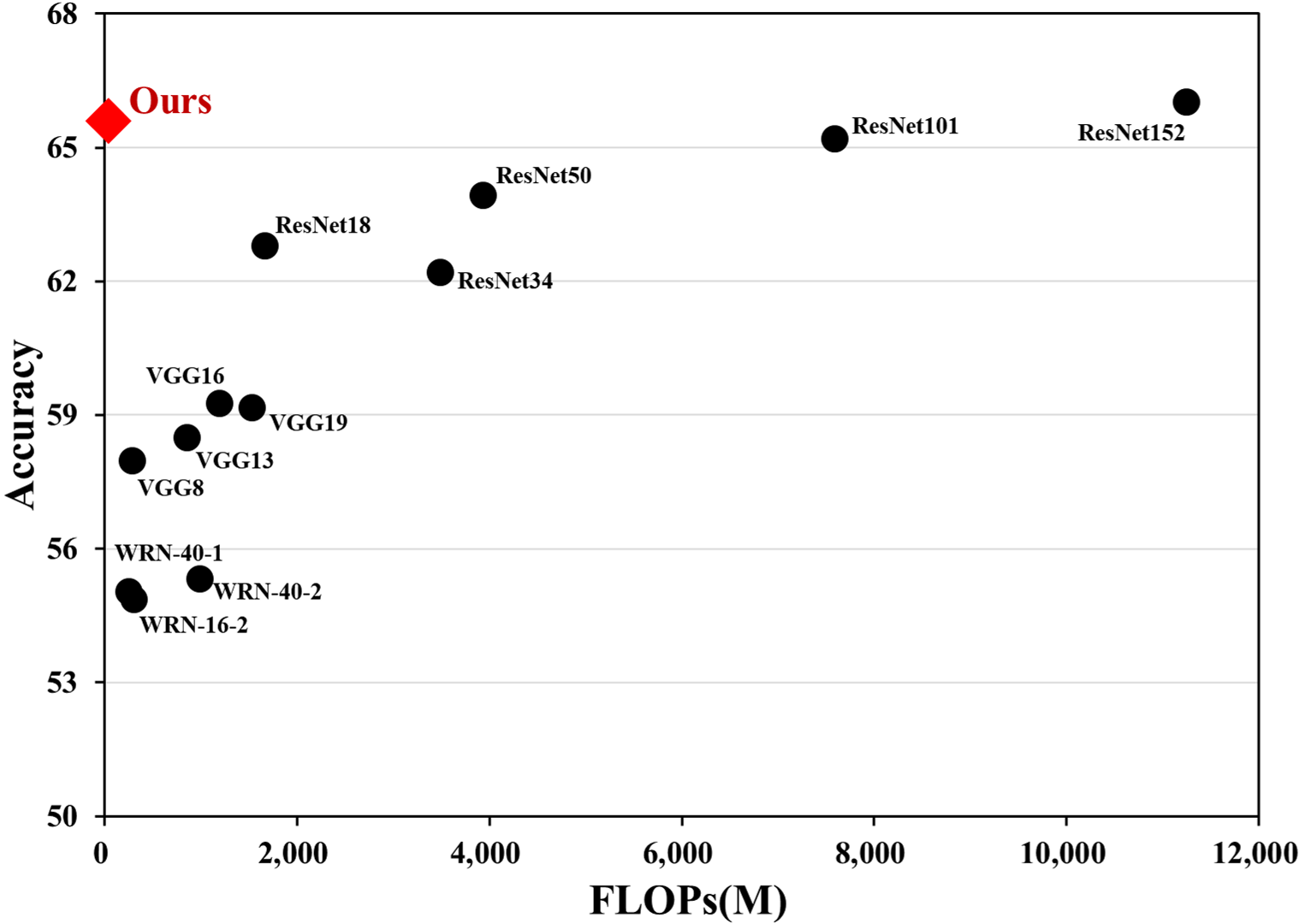}\\
	\vspace{-4mm}
	\caption{
		Performance comparison between ours and other models on CIFAR-100.  Smaller FLOPs represent more efficient models. Higher accuracy represents models have more excellent performance. 
	}
	\vspace{-7mm}
	\label{acc_param}
\end{figure} 

\noindent \textbf{Implementation:} For the comparative methods, the pretrained ResNet-34, ResNet50, ResNet101, ResNet152, VGG13, VGG16, VGG19, and WRN-40-2 are the teacher networks. Meanwhile, the ResNet18, VGG-8, WRN-16-2 and WRN40-1 are the student networks. For ours, the same pretrained network is employed as the CNN teacher. Graph head $f_{\star}^g(\cdot)$ is a deep neural network with $512$-$256$. The student network is a two layers GNN. We adopt the same graph normalization $\bm P=\phi(\bm A)$ in \cite{kipf2016semi} and adopt Mechanism \ref{batch_instance} to achieve inductive learning. Meanwhile, $s=50$ in Eq. (\ref{graph_solve}) and $\alpha=1$ in Eq. (\ref{student_batch}). The mini-batch gradient descent is employed to train both methods. The batch is set as $100$ and the epoch is $200$. The learning rate $\alpha$ is $0.01$. 


\subsection{Analysis of Experiments}
All models are run $10$ times on benchmark datasets and the mean value on CIFAR-100 and Mini-ImageNet are shown in Table \ref{Results_cifar100} and Table \ref{Results_miniimagenet}. The green up-arrow and the red down-arrow are higher performance and lower performance compared to KD, respectively. The other results are shown in the supplementary. Based on this, we can conclude that:
\begin{itemize}[leftmargin=10pt, topsep=0pt, itemsep=0pt]
	\item \textbf{Compared with other KDs:} Compared with the others, ours has achieved the highest accuracy on all datasets. Due to exploring the latent topological relationship, the proposed model can distinguish more complex images in Mini-ImageNet and the accuracy is far superior to others.
	\item \textbf{Compared with Teacher Network:} Notably, different from the KD and DKD distillation whose performance generally does outperform the teacher, ours can even obtain a higher performance than the teacher, especially on the complex Mini-ImageNet dataset. It is mainly caused by the distilled ``boosted'' GNN student who can not only learn the intra-sample representation generated from CNN but also explore the latent relationship among the samples.
	\item \textbf{Single GNN Student:} Equipped with the proposed graph head, the GNN student can be extended on non-graph datasets and obtain satisfactory performance on CIFAR-100 and Mini-ImageNet. Interestingly, compared with the CNN-based model, the ``boosted'' GNN student can bridge these two heterogeneous networks and achieve the best performance-efficient trade-off as shown in Fig. \ref{acc_param}.
\end{itemize}


\begin{figure}[t]
	\centering
	\vspace{-3mm}
	\subfigure[CIFAR100]{
		\label{cifar100}
		\includegraphics[scale=0.20]{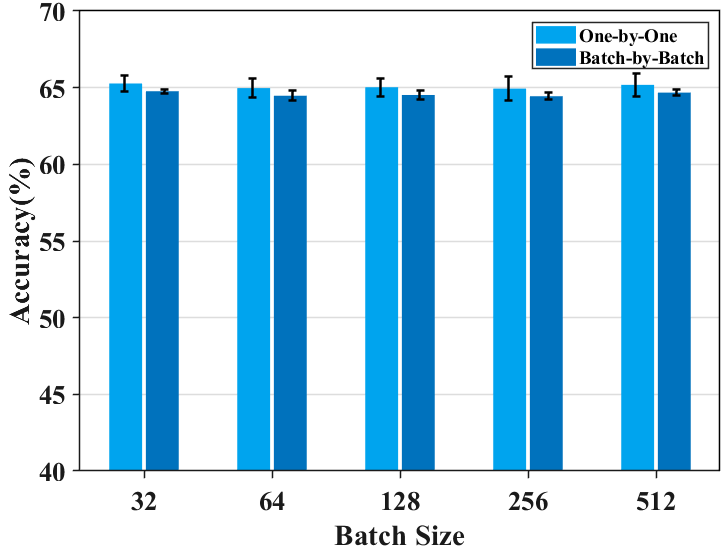}
	}
	\subfigure[Mini-ImageNet]{
		\label{mini-imagenet}
		\includegraphics[scale=0.20]{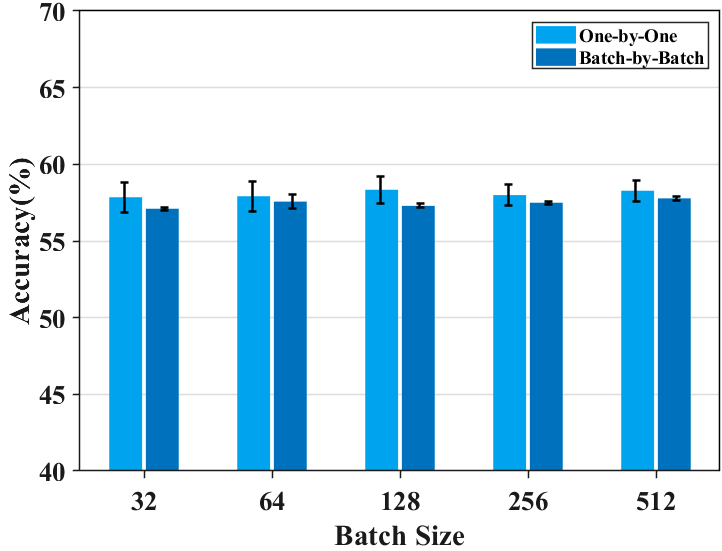}
	}
	\vspace{-5mm}
	\caption{Accuracy of CNN2GNN w.r.t the different batch size. Figure \ref{cifar100} and \ref{mini-imagenet} are the results on CIFAR-100 and Mini-ImageNet.}
	\label{bz_tune}
	\vspace{-5mm}	
\end{figure}

\subsection{Sensitivity Analysis}

In this part, we conduct the corresponding experiments to study the sensitivity of the sparsity $s$ in Eq. (\ref{graph_solve}) and the temperature $\tau$ in Eq. (\ref{student_batch}). The two architectures, ResNet34-18 and VGG13-8, are employed. From the results, the model is sensitive to the temperature $\tau$ and a higher temperature may bring higher performance, especially on CIFAR-100. It also means that the distillation between heterogeneous neural networks needs to keep a high temperature. 
Meanwhile, since Mini-Imagenet is more complex compared with CIFAR-100, a higher sparsity will bring more connectivity between the instances to improve the performance. 
Therefore, we can either fine-tune $\tau$ in $(2^4, 2^6]$ and simply set them as a median like $48$. For the complex datasets, $s$ can be set as higher.

\subsection{Ablation Study} \label{ablation}

We conduct an ablation study to evaluate how each part including Mechanism \ref{one_instance} (One-by-One), Mechanism \ref{batch_instance} (Batch-by-Batch), inner product-based GNN (InGNN), and Euclidean distance-based GNN (EucGNN) on CIFAR-100 and Mini-ImageNet. 
As shown in Table \ref{Ablation Study}, we can conclude that:

\textbf{Mechanism \ref{one_instance} vs Mechanism \ref{batch_instance}:} Compared with Batch-by-Batch, the time cost of One-by-One is much higher, which proves that the approximate-based strategy in Mechanism \ref{batch_instance} can significantly improve efficiency. Meanwhile, the accuracy of Batch-by-Batch is lower than One-by-One on CIFAR-100, which proves that this approximate-based strategy will sacrifice accuracy. As shown in Fig. \ref{bz_tune}, although the performance of the two mechanisms is stable with the batch size, One-by-One has a higher performance and standard deviation. Besides, since the proposed graph head will learn a reliable sparse graph after training, Mechanism \ref{batch_instance} can learn a approximated graph which correctly reflect the latent relationship among the test data. In a word, Mechanism \ref{batch_instance} can properly balance the accuracy and efficiency of inductive learning.

\textbf{Differentiable Sparse Graph vs Others:} One-by-One and Batch-by-Batch are both superior higher than InGNN and EucGNN on these two datasets. As shown in Table \ref{Results_cifar100}, the single ``boosted'' GNN student can also achieve higher accuracy than others. These two situations turn out that the designed sparse graph head can successfully learn a reliable graph structure by gradient descent and inductive learning. Furthermore, we notice that the Batch-by-Batch time cost is the fewest among these three models. Thus, the designed graph head can differentiable learns a sparse graph on non-graph data for inductive learning without adding the calculation cost.

\begin{table}[t]
	\renewcommand\arraystretch{1.2}
	\centering
	\vspace{-1mm}
	\caption{Ablation Study on CIFAR-100 and Mini-ImageNet}
	\vspace{-3mm}
	\label{Ablation Study}
	\scalebox{0.75}{
		\begin{tabular}{c|cc|cc}
			\toprule
			\multicolumn{1}{c|}{\multirow{2}{*}{Method}}                    & \multicolumn{2}{c|}{CIFAR-100}                    & \multicolumn{2}{c}{Mini-ImageNet}                  \\ \cline{2-5} 
			\multicolumn{1}{c|}{}                                            & \multicolumn{1}{c}{Accuracy(\%)} & \multicolumn{1}{c|}{Time(s)} & \multicolumn{1}{c}{Accuracy(\%)} & \multicolumn{1}{c}{Time(s)} \\ \midrule
			\multirow{1}{*}{InGNN} & 7.82                   & 2828.11                   & 6.17                   & 2567.47                   \\
			\multirow{1}{*}{EucGNN} & 6.15                  & \underline{2824.16}       & 5.82                   & \underline{2469.68}                   \\
			\multirow{1}{*}{One-by-One} & \textbf{59.77}        & 4999.39                  & \underline{48.85}                   & 3214.45                   \\
			\multirow{1}{*}{Batch-by-Batch} & \underline{58.57}     & \textbf{2797.42}          & \textbf{49.07}                  & \textbf{2307.49}                   \\
			\bottomrule
		\end{tabular}
	}
	\vspace{-4mm}
\end{table}	

\section{Conclusion}
In this paper, we propose a novel heterogeneous distillation network, CNN2GNN, to distill the knowledge from CNN to GNN, which successfully solves the concerned question of how to bridge CNN and GNN. Firstly, a graph head is designed to learn a sparse graph differentiable and inductively. It can extend the GNN on non-graph data and reduce the inference latency on large-scale datasets. Based on this, the knowledge extracted from a large CNN will be distilled into a small GNN. The distilled ``boosted'' GNN can inductively utilize a few neural layers to extract the deep intra-sample representation of a single instance and the topological relationship among the instances simultaneously. In experiment, the proposed network achieves excellent results.

\bibliographystyle{named}
\bibliography{egbib}

\end{document}